\documentclass[doublespacing]{elsart}
\usepackage{graphicx}
\usepackage{amssymb}

\usepackage{url}

\usepackage{natbib}

\usepackage{algorithmic}
\usepackage{algorithm}
\usepackage{scheme}

\newcounter{enumv}

\newcommand{\vdata}{\mathbf{x}}

\newcommand{\vmodel}{\mathbf{m}}
\newcommand{\vmodelfull}{\mathcal{M}}
\newcommand{\graph}{\mathcal{G}}
\newcommand{\vertices}{\mathcal{V}}
\newcommand{\edges}{\mathcal{E}}
\newcommand{\graphdist}{g}
\newcommand{\neigh}{h}
\newcommand{\kernel}{K}
\newcommand{\affect}{c}

\newcommand{\itermax}{L}
\newcommand{\cluster}{\mathcal{C}}
\newcommand{\diss}{d}
\newcommand{\inputspace}{\mathcal{X}}
\newcommand{\inputset}{\mathcal{D}}

\begin{document}
\sloppy

\begin{frontmatter}

% Title, authors and addresses

% use the thanksref command within \title, \author or \address for footnotes;
% use the corauthref command within \author for corresponding author footnotes;
% use the ead command for the email address,
% and the form \ead[url] for the home page:
% \title{Title\thanksref{label1}}
% \thanks[label1]{}
% \author{Name\corauthref{cor1}\thanksref{label2}}
% \ead{email address}
% \ead[url]{home page}
% \thanks[label2]{}
% \corauth[cor1]{}
% \address{Address\thanksref{label3}}
% \thanks[label3]{}

%\protect\footnote{Accept\'e par \emph{Neural Networks} avec modifications mineures  \`a r\'ealiser.}
\title{Fast Algorithm and Implementation of Dissimilarity Self-Organizing
  Maps}

% use optional labels to link authors explicitly to addresses:
% \author[label1,label2]{}
% \address[label1]{}
% \address[label2]{}

\author[LITA]{Brieuc Conan-Guez}, %
\ead{Brieuc.Conan-Guez@univ-metz.fr}%
\author[INRIA]{Fabrice Rossi\corauthref{cor}}, %
\ead{Fabrice.Rossi@inria.fr}%
and \author[INRIA]{A{\"\i}cha El Golli}
\ead{Aicha.ElGolli@inria.fr}

\address[LITA]{LITA EA3097, Universit\'e de Metz, Ile du Saulcy, F-57045
  Metz, France} 
\address[INRIA]{Projet AxIS, INRIA, Domaine de Voluceau, Rocquencourt,
  B.P. 105, 78153 Le Chesnay Cedex, France}

\corauth[cor]{Corresponding author.}
\end{frontmatter}
\thispagestyle{empty}
\newpage
\begin{frontmatter}
\begin{abstract}
In many real world applications, data cannot be accurately represented by
vectors. In those situations, one possible solution is to rely on
dissimilarity measures that enable sensible comparison between observations. 

Kohonen's Self-Organizing Map (SOM) has been adapted to data described only
through their dissimilarity matrix. This algorithm provides both non linear
projection and clustering of non vector data. Unfortunately, the algorithm
suffers from a high cost that makes it quite difficult to use with voluminous
data sets. In this paper, we propose a new algorithm that provides an important
reduction of the theoretical cost of the dissimilarity SOM without changing
its outcome (the results are exactly the same as the ones obtained with the
original algorithm). Moreover, we introduce implementation methods that result
in very short running times.

Improvements deduced from the theoretical cost model are validated on
simulated and real world data (a word list clustering problem). We also
demonstrate that the proposed implementation methods reduce by a
factor up to 3 the running time of the fast algorithm over a standard
implementation.
\end{abstract}

\begin{keyword}
Fast implementation \sep Self Organizing Map \sep Clustering \sep Non linear
projection \sep Unsupervised learning \sep Dissimilarity Data \sep Proximity
Data \sep Pairwise Data
% keywords here, in the form: keyword \sep keyword

% PACS codes here, in the form: \PACS code \sep code
%\PACS 
\end{keyword}
\end{frontmatter}
\thispagestyle{plain}

% main text

\section{Introduction}
The vast majority of currently available data analysis methods are based on a
vector model in which observations are described with a fixed number of real
values, i.e., by vectors from a fixed and finite dimensional vector space.
Unfortunately, many real world data depart strongly from this model. It is
quite common for instance to have variable size data. They are natural for
example in online handwriting recognition \citep{Bahlman04Online} where the
representation of a character drawn by the user can vary in length because of
the drawing conditions. Other data, such as texts for instance, are strongly
non numerical and have a complex internal structure: they are very difficult
to represent accurately in a vector space. While a lot of work has been done
to adapt classical data analysis methods to structured data such as tree and
graph \citetext{see e.g., \citealp{Hammer04Structured} for neural based
  unsupervised processing of structured data and also
  \citealp{HammerJain04NonStandardData,HammerVillmann05NonStandardMetric}}, as
well as to data with varying size, there is still a strong need for efficient
and flexible data analysis methods that can be applied to any type of data.

A way to design such methods is to rely on one to one comparison between
observations. It is in general possible to define a similarity or a
dissimilarity measure between arbitrary data, as long as comparing them is
meaningful. In general, data analysis algorithms based solely on
(dis)similarities between observations are more complex than their vector
counterparts, but they are universal and can therefore be applied to any kind
of data. Moreover, they allow one to rely on specific (dis)similarities
constructed by experts rather than on a vector representation
of the data that induces in general unwanted distortion in the observations.

Many algorithms have been adapted to use solely dissimilarities between data.
In the clustering field, the k-means algorithm \citep{MacQueenKMeans1967} has
been adapted to dissimilarity data under the name of Partitioning Around
Medoids \citep[PAM,][]{KaufmanRousseeuw1987PAM}. More recently, approaches
based on deterministic annealing have been used to propose another class of
extensions of the k-means principle
\citep[see][]{BuhmannHofmann1994ICPR,HofmannBuhmann1995ICANN,HofmannBuhmann1997TPAMI}.
Following the path taken for the k-means, several adaptation of Kohonen's
Self-Organizing Map \citep[SOM,][]{KohonenSOM1995} to dissimilarity data have
been proposed. \citet{AmbroiseGovaert1996} proposed a probabilistic
formulation of the SOM that can be used directly for dissimilarity data.
Deterministic annealing schemes have been also used for the SOM
\citep{GraepelEtAl1998GSOM,GraepelObermayer1999DSOM,SeoObermayer2004}.  In the
present paper, we focus on an adaptation proposed in
\citep{KohonenSomervuo1998Symbol,KohonenSomervuo2002NonVectorial}, where it
was applied successfully to a protein sequence clustering and visualization
problem, as well as to string clustering problems. This generalization is
called the Dissimilarity SOM (DSOM, also known as the median SOM), and can be
considered as a SOM formulation of the PAM method. Variants of the DSOM were
applied to temperature time series \citep{ElGolliConanGuezRossi04JSDA},
spectrometric data \citep{ElGolliConanGuezRossiIFCS2004SomDiss} and web usage
data \citep{RossiEtAlESANN2005WUMSOM}. 

A major drawback of the DSOM is that its running time can be very high,
especially when compared to the standard vector SOM. It is well known that the
SOM algorithm behaves linearly with the number of input data \citep[see,
e.g.][]{KohonenSOM1995}. On the contrary, the DSOM behaves quadratically with
this number (see Section \ref{subsectionCosts}). We propose in this paper
several modifications of the basic algorithm that allow a much faster
implementation. The quadratic nature of the algorithm cannot be avoided,
essentially because dissimilarity data are intrinsically described by a
quadratic number of one to one dissimilarities. Nevertheless, the standard
DSOM algorithm cost is proportional to $N^2M+NM^2$, where $N$ is the number of
observations and $M$ the number of clusters that the algorithm has to produce,
whereas our modifications lead to a cost proportional to $N^2+NM^2$. Moreover,
a specific implementation strategy reduces the actual computation
burden even more. An important property of all our modifications is that the
obtained algorithm produces \textbf{exactly} the same results as the standard
DSOM algorithm.

The paper is organized as follows. In section \ref{sectionDSOM} we recall the
SOM adaptation to dissimilarity data and obtain the theoretical cost of the
DSOM. In section \ref{sectionFastDSOM}, we describe our proposed new algorithm
as well as the implementation techniques that decrease its running time in
practice. Finally we evaluate the algorithm in section \ref{subsectionExp}.
This evaluation validates the theoretical cost model and shows that the
implementation methods reduce the running time. The evaluation is conducted on
simulated data and on real world data (a word list clustering problem).

\section{Self-Organizing Maps for dissimilarity data}\label{sectionDSOM}
\subsection{A batch SOM for dissimilarity data}
We recall in this section the DSOM principle as proposed in
\citep{KohonenSomervuo1998Symbol,KohonenSomervuo2002NonVectorial}. Let us
consider $N$ input data from an arbitrary input space $\inputspace$,
$(\vdata_i)_{1\leq i\leq N}$. The set of those $N$ data is denoted
$\inputset$. The only available information on the data set is the
dissimilarities between its elements: the dissimilarity between $\vdata_i$ and
$\vdata_k$ is denoted $\diss(\vdata_i,\vdata_k)$. We assume standard
dissimilarity behavior for $\diss$, that is: $\diss$ is
symmetric, positive and $\diss(\vdata_i,\vdata_i)= 0$.

As the standard SOM, the DSOM maps input data from an input space to a low
dimensional organized set of $M$ models (or neurons) which are numbered from
$1$ to $M$, and arranged via a prior structure (a grid in general). Model $j$
is associated to an element of $\inputset$, its prototype, denoted $\vmodel_j$
(therefore, for each $j$, there is $i$ that depends on $j$, such that
$\vmodel_j=\vdata_i$): this is the first difference with the standard SOM in
which prototypes can take arbitrary values in the input space.

The prior structure on the models is represented by an undirected graph
$\graph=(\vertices,\edges)$ whose vertices are model numbers (i.e.,
$\vertices=\{1,\ldots,M\}$). We denote $\graphdist(j,k)$ the length of the
shortest path in $\graph$ from $j$ to $k$. Given a kernel like function
$\kernel$, that is a decreasing function from $\Rset^+$ to $\Rset^+$, with
$\kernel(0)=1$ and $\lim_{s\rightarrow\infty}\kernel(s)=0$, the neighborhood
relationship between models $j$ and $k$, $\neigh(j,k)$, is defined by
$\neigh(j,k)=\kernel(\graphdist(j,k))$. As for the standard SOM, the kernel is
modified during training: at the beginning, the neighborhood is very broad to
allow organization to take place. The kernel sharpens during training and
models become more and more adapted to a small subset of the data.

Given those information, the Dissimilarity SOM algorithm can be defined (see
Algorithm \ref{algoDSOM}). 

\begin{algorithm}
\caption{The Dissimilarity  SOM}\label{algoDSOM}
  \begin{algorithmic}[1]
\STATE choose initial values for
    $\vmodelfull^0=(\vmodel_1^0,\ldots,\vmodel_M^0)$ \COMMENT{Initialization
      phase}
\FOR{$l=1$ to $\itermax$}
  
  \FORALL[Template for the affectation phase]{$i\in \{1,\ldots,N\}$}
  \STATE compute  
    \begin{equation}\label{eqAffectDSOM}
      \affect^l(i)=\arg\min_{j\in
        \{1,\ldots,M\}}\diss(\vdata_i,\vmodel^{l-1}_j). 
    \end{equation}
 \ENDFOR
  \FORALL[Template for the representation phase]{$j\in \{1,\ldots,M\}$}
   \STATE compute 
    \begin{equation}
      \label{eqRepreDSOM}
      \vmodel_j^l=\arg\min_{\vmodel\in\inputset}\sum_{i=1}^N\neigh^l(\affect^l(i),j)\diss(\vdata_i,\vmodel).
    \end{equation}
  \ENDFOR
\ENDFOR
  \end{algorithmic}
\end{algorithm}
While some initialization techniques proposed for the standard SOM can be
extended to the case of dissimilarity data (see
\citep{KohonenSomervuo1998Symbol}), we use in this article a simple random
choice: $\vmodelfull^0$ is a random subset of the data set.

After initialization, the algorithm runs for $\itermax$ epochs. One epoch
consists in an \emph{affectation phase}, in which each input is associated to
a model, followed by a \emph{representation phase} in which the prototype of
each model is updated. The DSOM is therefore modelled after the batch version
of the SOM. As mentioned above, the neighborhood relationship depends on $l$:
at epoch $l$, we use $\neigh^l$ (see Equation \ref{eqRepreDSOM}).

At the end of the algorithm, an additional affectation phase can be done to
calculate $\affect^{l+1}(i)$ for all $i$ and to define $M$ clusters
$\cluster^{l+1}_1,\ldots,\cluster^{l+1}_M$ with $\cluster^{l+1}_j=\{1\leq
i\leq N\mid \affect^{l+1}(i)=j\}$.

It should be noted that in practice, as pointed out in
\citep{KohonenSomervuo1998Symbol}, the simple affectation phase of Equation
\ref{eqAffectDSOM} induces some difficulties: for certain types of
dissimilarity measures, the optimization problem of Equation
\ref{eqAffectDSOM} has many solutions. A tie breaking rule should be used. In
this paper, we use the affectation method proposed in
\citep{KohonenSomervuo1998Symbol}. In short, it consists in using a growing
neighborhood around each neuron to build an affinity of a given observation to
the neuron. Details can be found in \citep{KohonenSomervuo1998Symbol}. Other
tie-breaking methods have been proposed, for instance in
\citep{ElGolliConanGuezRossiIFCS2004SomDiss,ElGolliConanGuezRossi04JSDA}. They
give similar results and have the same worst case complexity. However, the
method of \citep{KohonenSomervuo1998Symbol} has a smaller best case
complexity.

Algorithm \ref{algoDSOM} provides a general template. In the rest of this
paper, we provide mostly partial algorithms (called Schemes) that fill the
missing parts of Algorithm \ref{algoDSOM}. 

\subsection{Algorithmic cost of the DSOM algorithm}\label{subsectionCosts}
For one epoch of the DSOM, there is one affectation phase, followed by one
representation phase. The affectation phase proposed in
\citep{KohonenSomervuo1998Symbol} has a worst case complexity of $M^2$ for one
observation and induces therefore a total cost of $O(NM^2)$ (the best case
complexity is $O(NM)$ when the optimization problem of Equation
\ref{eqAffectDSOM} has only one solution for each observation).

The major drawback of the DSOM algorithm is the cost induced by the
representation phase: there is no closed formula for the optimization used in
Equation \ref{eqRepreDSOM} and some brute force approach must be used. The
simple solution used in
\citep{KohonenSomervuo1998Symbol,KohonenSomervuo2002NonVectorial,ElGolliConanGuezRossiIFCS2004SomDiss,ElGolliConanGuezRossi04JSDA}
consists in an elementary search procedure: each possible value for
$\vmodel^l_j$ is tested (among $N$ possibilities) by computing the sum
$\sum_{i=1}^N\neigh^l(\affect^l(i),j)\diss(\vdata_i,\vmodel)$. This method is
called the brute force Scheme (implementation is obvious and therefore is not
given here). The calculation of one sum costs $O(N)$ and there are $N$ sums to
test, for a total cost of $O(N^2)$. For the whole representation phase, the
total cost is therefore $O(N^2M)$.

The total cost for the DSOM for one epoch (with the brute force Scheme and
using the affectation rule of \citep{KohonenSomervuo1998Symbol}) is therefore
$O(NM^2+N^2M)$. In general $M$ is much smaller than $N$ and therefore, the
representation phase clearly dominates this cost.

\section{A fast implementation}\label{sectionFastDSOM}
\subsection{Related works}
A lot of work has been done in order to optimize clustering algorithms in
terms of running time. However, most of those works have two limitations: they
are restricted to vector data and they produce different results from the
original algorithms \citetext{see e.g. \citealp{KohonenWEBSOM2000} for this
  type of optimizations of the SOM algorithm and
  \citealp{KohonenSomervuo2002NonVectorial} for the DSOM}. A review and a
comparison of optimized clustering algorithms for dissimilarity data are given
in \citep{WeietAl2003}: the four distinct algorithms give different results on
the same data set. While the algorithms try to solve the same problem as the
PAM method \citep{KaufmanRousseeuw1987PAM}, they also give different results
from PAM itself. On the contrary, in this paper, we focus on modifying the
DSOM algorithm without modifying its results. We will therefore avoid
optimization techniques similar to the ones reviewed in \citep{WeietAl2003},
for instance the sampling method used in
\citep{KohonenSomervuo2002NonVectorial}.

\subsection{Partial sums}
The structure of the optimization problem of Equation \ref{eqRepreDSOM} can be
leverage to provide a major improvement in complexity. At epoch $l$ and for
each model $j$, the goal is to find for which $k$, $S^l(j,k)$ is minimal,
where $S^l(j,k)$ is given by
\begin{equation}
  \label{eqSum}
S^l(j,k)=  \sum_{i=1}^N\neigh^l(\affect^l(i),j)\diss(\vdata_i,\vdata_k).
\end{equation}
Those sums can be rewritten as follows
\begin{equation}
  \label{eqSumWithPartial}
  S^l(j,k)=\sum_{u=1}^M \neigh^l(u,j)D^l(u,k),
\end{equation}
with 
\begin{equation}
  \label{eqPartialDist}
D^l(u,k)= \sum_{i\in \cluster^l_u}\diss(\vdata_i,\vdata_k).
\end{equation}
There are $MN$ $D^l(u,k)$ values, which can be calculated has a whole
with $O(N^2)$ operations: calculating one $D^l(u,k)$ costs
$O(|\cluster^l_u|)$. Then calculating all the $D^l(u,k)$ for a fixed $u$
costs $O(N|\cluster^l_u|)$. As $\sum_{u=1}^M|\cluster^l_u|=N$, the total
calculation cost is $O(N^2)$.

\begin{scheme}
\caption{An efficient scheme for the representation phase}\label{schemePartialSum}
  \begin{algorithmic}[1]
\FORALL[Calculation of the $D^l(u,k)$]{$u\in\{1,\ldots,M\}$}
  \FORALL{$k\in\{1,\ldots,N\}$}
    \STATE $D^l(u,k)\leftarrow 0$
    \FORALL{$i\in \cluster^l_u$}
      \STATE $D^l(u,k)\leftarrow D^l(u,k) + \diss(\vdata_i,\vdata_k)$
    \ENDFOR
  \ENDFOR
\ENDFOR
\FORALL[Representation phase]{$j\in\{1,\ldots,M\}$}
  \STATE $\delta\leftarrow \infty$
  \FORALL[outer loop]{$k\in\{1,\ldots,N\}$}
    \STATE $\delta_k\leftarrow \neigh^l(1,j)D^l(1,k)$
    \FORALL[inner loop]{$u\in\{1,\ldots,M\}$}
      \STATE $\delta_k\leftarrow \delta_k + \neigh^l(u,j)D^l(u,k)$
    \ENDFOR
    \IF{$\delta_k<\delta$}
      \STATE $\delta\leftarrow\delta_k$
      \STATE $\vmodel^l_j\leftarrow\vdata_k$ 
    \ENDIF
  \ENDFOR
\ENDFOR
  \end{algorithmic}
\end{scheme}

The calculation of one $S^l(j,k)$ using pre-calculated values of the
$D^l(u,k)$ can therefore be done in $O(M)$ operations. The representation
phase for model $j$ needs the values of $S^l(j,k)$ for all $k$ and the total
cost is therefore $O(NM)$. As the $D^l(u,k)$ can be calculated once for all
models, the total cost of the representation phase is $O(N^2+NM^2)$, whereas
it was $O(N^2M)$ for the brute force scheme. As $M<N$ in almost all
situations, this approach reduces the cost of the DSOM. Scheme
\ref{schemePartialSum} gives the proposed solution. 

\subsection{Early stopping}\label{sectionEarlyStopping}
While Scheme \ref{schemePartialSum} is much more efficient in practice than
the brute force Scheme, additional optimizations are available. The
simplest one consists in using an early stopping strategy for the inner loop
(line 13 of Scheme \ref{schemePartialSum}): the idea is to move into the loop
an adapted version of the test that starts on line 16 (of Scheme
\ref{schemePartialSum}). It is pointless to calculate the exact value of
$S^l(j,k)$ (i.e., $\delta_k$ in the algorithm) if we already know for sure
that this value is higher than a previously calculated one. This optimization
does not reduce the worst case complexity of the algorithm and has an overhead
as it involves an additional comparison in the inner loop. It will therefore
be only useful when $M$ is high and when the data induce frequent early
stopping. In order to favor early stopping, both the inner loop and the outer
loop should be ordered. During the inner loop, the best order would be to sum
first high values of $\neigh^l(u,j)D^l(u,k)$ so as to increase $\delta_k$ as
fast as possible. For the outer loop, the best order would be to start with
low values of $S^l(j,k)$ (i.e., with good candidates for the prototype of
model $j$): a small value of $\delta$ will stop inner loops earlier than a
high value.

In practice however, computing optimal evaluation orders will induce an
unacceptable overhead. We rely therefore on ``good'' evaluation orders induced
by the DSOM itself. The standard definition of $\neigh^l$ implies that
$\neigh^l(u,j)$ is small when model $u$ and model $j$ are far away in the
graph. It is therefore reasonable to order the inner loop on $u$ in decreasing
order with respect to $\neigh^l(u,j)$, that is in increasing order with
respect to the graph distance between $u$ and $j$, $\graphdist(u,j)$.

For the outer loop, we leverage the organization properties of the DSOM:
observations are affected to the cluster whose prototype is close to them.
Therefore, the \emph{a priori} quality of an observation $\vdata_k$ as a
prototype for model $j$ is roughly the inverse of the distance between $j$ and
the cluster of $\vdata_k$ in the prior structure. Moreover, the prototype
obtained during the previous epoch should also be a very good candidate for
the current epoch. 

To define precisely evaluation orders, let us choose, for all
$j\in\{1,\ldots,M\}$, $\zeta_j$, a permutation of $\{1,\ldots,M\}$ such that
for all $u\in \{1,\ldots,M-1\}$, $\graphdist(\zeta_j(u),j)\geq
\graphdist(\zeta_j(u+1),j)$. Scheme \ref{schemePartialSumEarlyStopOrderNeigh}
is constructed with this permutation. It should be noted that this Scheme
gives exactly the same numerical results as Scheme \ref{schemePartialSum}.

\begin{scheme}
\caption{Neighborhood based ordered representation phase}\label{schemePartialSumEarlyStopOrderNeigh}
  \begin{algorithmic}[1]
\STATE Calculation of the $D^l(u,k)$ \COMMENT{See lines 1--8 of Scheme \ref{schemePartialSum}}
\FORALL[Representation phase]{$j\in\{1,\ldots,M\}$}
  \STATE $\delta\leftarrow \infty$
  \FOR[outer cluster loop]{$v=1$ to $M$}
    \FORALL[outer candidate loop]{$k\in \cluster^l_{\zeta_j(v)}$}
      \STATE $\delta_k\leftarrow 0$
      \FOR[inner loop]{$u=1$ to $M$}
        \STATE $\delta_k\leftarrow \delta_k + \neigh^l(\zeta_j(u),j)D^l(\zeta_j(u),k)$
        \IF[early stopping]{$\delta_k>\delta$}
        \STATE \textbf{break} inner loop
        \ENDIF
      \ENDFOR
      \IF{$\delta_k<\delta$}
        \STATE $\delta\leftarrow\delta_k$
        \STATE $\vmodel^l_j\leftarrow\vdata_{k}$ 
      \ENDIF
    \ENDFOR
  \ENDFOR
  \STATE put $\vmodel^l_j$ at the first position in its cluster
\ENDFOR
  \end{algorithmic}
\end{scheme}

Line 19 prepares the next epoch by moving the prototype at the first position
in its cluster: this prototype will be tested first in the next epoch.
Except for this special case, we don't use any specific order
inside each cluster.

\subsection{Reusing earlier values}\label{sectionMemory}
Another source of optimizations comes from the iterative nature of the DSOM
algorithm: when the DSOM algorithm proceeds, clusters tend to stabilize and
$D^{l+1}(u,k)$ will be equal to $D^{l}(u,k)$ for many pairs $(u,k)$. 

This stabilization property can be used to reduce the cost of the first phase
of Scheme \ref{schemePartialSum}. During the affectation phase, we simply have
to monitor whether the clusters are modified. If
$\cluster^{l-1}_u=\cluster^l_u$, then for all $k\in\{1,\ldots,N\}$,
$D^{l-1}(u,k)=D^l(u,k)$. This method has a very low overhead: it only adds a
few additional tests in the affectation phase ($O(N)$ additional operations)
and in the pre-calculation phase ($O(M)$ additional operations). However this
block update method has a very coarse grain. Indeed, a full calculation of
$D^l(u,k)$ for two values of $u$ (i.e., two clusters) can be triggered by the
modification of the cluster of an unique observation. It is therefore tempting
to look for a finer grain solution. Let us consider the case where clusters
don't change between epoch $l-1$ and epoch $l$, except for one observation,
$\vdata_i$. More precisely, we have $\affect^{l-1}(k)=\affect^l(k)$ for all
$k\neq i$. Then for all $u$ different from $\affect^{l-1}(i)$ and
$\affect^l(i)$, $D^l(u,k)=D^{l-1}(u,k)$ (for all $k$). Moreover, it appears
clearly from Equation \ref{eqPartialDist}, that for all $k$
\begin{eqnarray}\label{eqUpdatePrototype}
D^l(\affect^{l-1}(i),k)&=&D^{l-1}(\affect^{l-1}(i),k)-\diss(\vdata_i,\vdata_k)\\
D^l(\affect^l(i),k)&=&D^{l-1}(\affect^l(i),k)+\diss(\vdata_i,\vdata_k)
\end{eqnarray}
Applying those updating formulae induces $2N$ additions and $N$ affectations
(loop counter is not taken into account). If several observations are moving
from their ``old'' cluster to a new one, updating operations can be performed
for each of them. In the extreme case where all observations have modified
clusters, the total number of additions would be $2N^2$ (associated to $N^2$
affectations). The pre-calculation phase of algorithm \ref{schemePartialSum}
has a smaller cost ($N^2$ additions and $N^2$ affectations). This means that
below approximately $\frac{N}{2}$ cluster modifications, the update approach
is more efficient than the full calculation approach for the $D^l(u,k)$
sums. 

\begin{algorithm}
\baselineskip 1.4em
  \caption{DSOM with memory}\label{algoDSOMSmartMemory}
  \begin{algorithmic}[1]
    \STATE Initialization \COMMENT{See line 1 of Algorithm \ref{algoDSOM}}
    \STATE $c^0(i)\leftarrow -1$ for all $i\in\{1,\ldots,N\}$ \COMMENT{this
      value triggers a full calculation of the $D^l(u,k)$ during the first epoch}
    \FOR{$l=1$ to $\itermax$}
       \STATE $v_u \leftarrow true$ for all $u\in\{1,\ldots,M\}$
       \COMMENT{Clusters have not changed yet}
       \STATE $nb\_switch \leftarrow 0$
       \FORALL[Affectation phase]{$i\in \{1,\ldots,N\}$}
          \STATE compute $\affect^l(i)$ with the method of \citep{KohonenSomervuo1998Symbol}
          \COMMENT{Any other affectation method can be used}
          \IF{$\affect^l(i)\neq \affect^{l-1}(i)$}
            \STATE $nb\_switch \leftarrow nb\_switch + 1$
            \STATE $v_{\affect^{l-1}(i)}\leftarrow false$ \COMMENT{cluster
              $\affect^{l-1}(i)$ has been modified}
            \STATE $v_{\affect^l(i)}\leftarrow false$ \COMMENT{cluster
              $\affect^{l}(i)$ has been modified}
          \ENDIF
       \ENDFOR
       \IF[Block update]{$nb\_switch\geq N/ratio$}
         \FORALL[Calculation of the $D^l(u,k)$]{$u\in\{1,\ldots,M\}$}
           \IF[New values must be calculated]{$v_u$ is false}
             \FORALL{$k\in \{1,\ldots,N\}$}
               \STATE $D^l(u,k)\leftarrow \sum_{i\in \cluster^l_u}\diss(\vdata_i,\vdata_k)$
             \ENDFOR
           \ELSE[Old values can be reused]
             \FORALL{$k\in \{1,\ldots,N\}$}
               \STATE $D^l(u,k)\leftarrow D^{l-1}(u,k)$
             \ENDFOR
           \ENDIF
         \ENDFOR
       \ELSE[Individual update]
          \FORALL{$i\in \{1,\ldots,N\}$}
            \IF{$\affect^l(i)\neq \affect^{l-1}(i)$}
              \FORALL{$k\in \{1,\ldots,N\}$}
                 \STATE $D^l(\affect^{l-1}(i),k)\leftarrow D^l(\affect^{l-1}(i),k) - \diss(\vdata_i,\vdata_k)$
                 \STATE $D^l(\affect^{l}(i),k)\leftarrow D^l(\affect^{l}(i),k) + \diss(\vdata_i,\vdata_k)$
              \ENDFOR
            \ENDIF
          \ENDFOR
       \ENDIF  
       \STATE Representation phase \COMMENT{See Schemes \ref{schemePartialSum}
         and \ref{schemePartialSumEarlyStopOrderNeigh}}
    \ENDFOR
  \end{algorithmic}  
\end{algorithm}

In order to benefit from both approaches, we use a hybrid algorithm
(Algorithm \ref{algoDSOMSmartMemory}). This algorithm chooses dynamically the
update method by counting the number of observations for which the
affectation result has changed. If this number is above a specified threshold
proportional to $N$, the block update method is used. If the number fails
below the threshold, the fine grain method is used.

\section{Evaluation of the proposed optimizations}\label{subsectionExp}
\subsection{Methodology}
The algorithms have been implemented in Java and tested with the runtime
provided by Sun (JDK 1.5, build \url{1.5.0_04-b05}). Programs have been
studied on a workstation equipped with a 3.00 GHz Pentium IV (Hyperthreaded)
with 512Mo of main memory, running the GNU/Linux operating system (kernel
version 2.6.11).  The Java runtime was set in server mode in order to activate
complex code optimization and to reduce Java overheads. For each algorithm,
the Java Virtual Machine (JVM) is started and the dissimilarity matrix is
loaded. Then, the algorithm is run to completion once. The timing of this run
is not taken into account as the JVM implements just in time compilation.
After completion of this first run and in the same JVM, ten subsequent
executions of the algorithm are done and their running time are averaged. The
reported figures are the corresponding average running time (on an otherwise
idle workstation) or ratio between reference running time and studied running
time. We do not report the standard deviation of the running times as it is
very small compared to the mean (the ratio between the standard deviation and
the mean is smaller than $1.52\,10^{-3}$ in 90\% of the experiments, with a
maximum value of $7.85\,10^{-2}$). This experimental setting was used in order
to minimize the influence of the operating system and of the implementation
language.

\subsection{Algorithms}
We have proposed several algorithms and several schemes for the affectation
phase. We have decided to test the combinations given in Table
\ref{tableAlgorithms}. We always used the affectation method of
\citep{KohonenSomervuo1998Symbol}.

\begin{table}[htbp]
  \centering
  \begin{tabular}{|l|c|c|}\hline
    Name & Algorithm & Representation Scheme \\\hline
    Brute force DSOM & \ref{algoDSOM} & brute force\\\hline
    Partial sum DSOM & \ref{algoDSOM} & \ref{schemePartialSum}\\\hline
    Early stopping DSOM & \ref{algoDSOM} &
    \ref{schemePartialSumEarlyStopOrderNeigh} \\\hline
    DSOM with memory & \ref{algoDSOMSmartMemory} & \ref{schemePartialSum}\\\hline
    Fast DSOM &\ref{algoDSOMSmartMemory}  & \ref{schemePartialSumEarlyStopOrderNeigh}\\\hline
  \end{tabular}
  \caption{Evaluated algorithms}
  \label{tableAlgorithms}
\end{table}

\subsection{Artificial data}
\subsubsection{Data and reference performances}\label{subsubReference}
The proposed optimized algorithms have been evaluated on a simple
benchmark. It consists in a set of $N$ vectors in $\Rset^2$ chosen randomly
and uniformly in the unit square. A DSOM with a hexagonal grid with
$M=m\times m$ models is applied to those data considered with the square
euclidean metric. We always used $\itermax=100$ epochs and a Gaussian
kernel for the neighborhood function. 

We report first some reference performances obtained with the brute force
DSOM (Algorithm \ref{algoDSOM} with the brute force Scheme). We have
tested five values for $N$ the number of observations, $500$, $1\,000$,
$1\,500$, $2\,000$ and $3\,000$. We tested four different sizes for the grid,
$M=49=7\times 7$, $M=100=10\times 10$, $M=225=15\times 15$ and $M=400=20\times
20$. To avoid too small clusters, high values of $M$ were used only with high
values of $N$. We report those reference performances in seconds in Table
\ref{tableBruteForce} (empty cells correspond to meaningless situations where
$M$ is too high relatively to $N$).

\begin{table}[htbp]
  \centering
  \begin{tabular}{|p{10em}|c|c|c|c|c|}\hline
$N$ (data size)\newline 
$M$ (number of models) & 500 & $1\,000$ & $1\,500$ & $2\,000$ & $3\,000$ \\\hline
$49=7\times 7$         & 11.4 &  53.5 &  135.4 &  261.6  &  865.3\\\hline 
$100=10\times 10$      & 24.7 & 115   &  283.4 &  557    & 1757.0\\\hline 
$225=15\times 15$      &      & 313.7 &  806.6 & 1594.8  & 4455.5\\\hline 
$400=20\times 20$      &      &       & 1336.9 & 2525.2  & 7151.8\\\hline 
  \end{tabular}
  \caption{Running time in seconds of the brute force DSOM algorithm}
  \label{tableBruteForce}
\end{table}

\begin{figure}[htbp]
  \centering
  \includegraphics[angle=270,width=\textwidth]{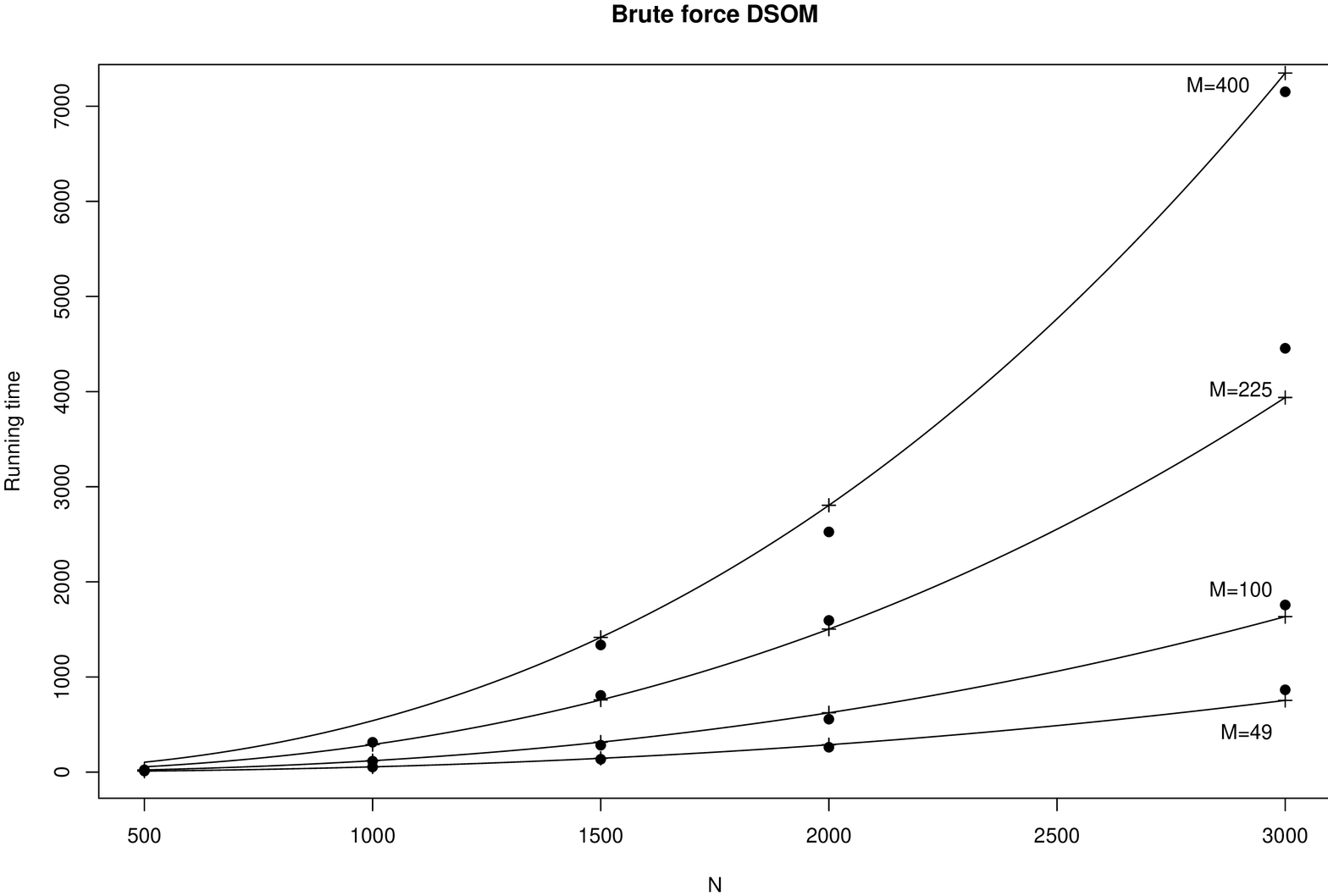}
  \caption{Running time for the brute force DSOM: solid circles are actual measurements, while plus signs and lines are estimation from the log-regression model}
  \label{figureBruteForceDSOM}
\end{figure}

The obtained values are compatible with the cost model. A least square
regression model for the running time $T$ of the form $\alpha N^2M$ is quite
accurate (the normalized mean square error, NMSE, i.e. the mean square error
of the model divided by the variance of $T$, is smaller than $0.016$).
However, because of the large differences between running times, the model is
dominated by the high values and is not very accurate for small values. A
simple logarithmic model ($\log T=\alpha\log N+\beta \log M+\gamma$) gives a
more accurate prediction for smaller values (the NMSE is smaller than
$0.0072$). In this case, $\alpha\simeq 2.37$ and $\beta\simeq 1.08$ (see
Figure \ref{figureBruteForceDSOM}). The real complexity is therefore growing
quicker than $N^2M$. This is a consequence of the hierarchical structure of
the memory of modern computers (a slow main memory associated to several
levels of faster cache memory). As the dissimilarity matrix does not fit into
the cache memory when $N$ is big, the calculation relies on the main memory,
which is slower than the cache. When $N$ is small, the computation model used
to derive the $N^2M$ cost is valid.  When $N$ is big enough, the model is too
simplistic and real performances are worse than expected.

\begin{figure}[htbp]
  \centering
  \includegraphics[angle=270,width=\textwidth]{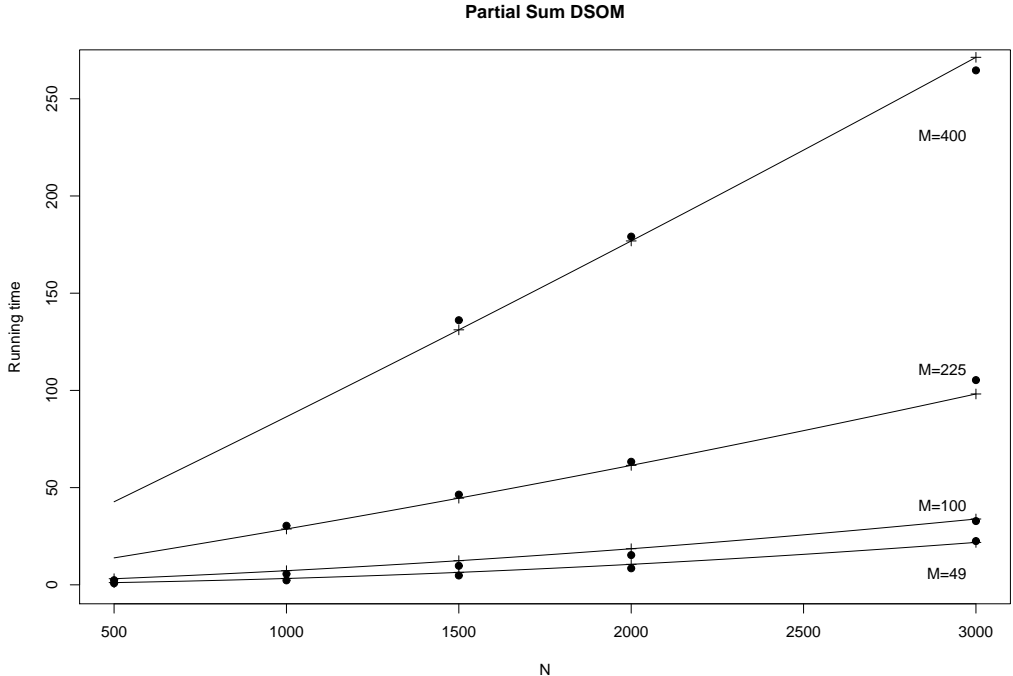}
  \caption{Running time for the partial sum DSOM: solid circles are actual measurements, while plus signs and lines are estimation from the regression model}
  \label{figurePartialSumDSOM}
\end{figure}

The other reference performances correspond to the partial sum DSOM (Algorithm
\ref{algoDSOM} with Scheme \ref{schemePartialSum}) and are summarized in Table
\ref{tablePartialSum}. Improvements over the brute force DSOM are quite
impressive. The running time $T$ is correctly modeled by $\delta N^2 +\tau
NM^2$ (the NMSE is smaller than $0.002$, see Figure
\ref{figurePartialSumDSOM}), where the ratio between $\delta$ and $\tau$ is
approximately $3.8$ (this version of the DSOM does not suffer too much from
the hierarchical structure of the memory).  According to the $N^\alpha
M^\beta$ model established above for the brute force algorithm, the ratio
between the running times should be proportional to $\frac{N^\alpha
  M^\beta}{3.8N^2 + NM^2}$, something that is verified easily on the data.

\begin{table}[htbp]
  \centering
  \begin{tabular}{|p{10em}|c|c|c|c|c|}\hline
$N$ (data size)\newline 
$M$ (number of models) & 500 & $1\,000$ & $1\,500$ & $2\,000$ & $3\,000$ \\\hline
$49=7\times 7$         & 0.8 &  2.3 &   4.8 &   8.5 &  22.5\\\hline 
$100=10\times 10$      & 2.4 &  5.6 &   9.8 &  15.3 &  32.8\\\hline 
$225=15\times 15$      &     & 30.4 &  46.4 &  63.3 & 105.3\\\hline 
$400=20\times 20$      &     &      & 136.1 & 179.1 & 264.6\\\hline 
  \end{tabular}
  \caption{Running time of the partial sum DSOM algorithm}
  \label{tablePartialSum}
\end{table}

Experiments with a bigger number of models show as expected less improvement
over the brute force DSOM algorithm, simply because the cost of the
representation part in Scheme \ref{schemePartialSum}, $O(NM^2)$, has a more
important role when $M$ increases and the algorithm is clearly behaving
quadratically with $M$ for a fixed value of $N$.

Those reference simulations show that the theoretical cost model is accurate
enough to predict the very important speed up. They also show that the
representation Scheme \ref{schemePartialSum}, based on partial sum, should
replace the brute force Scheme, in all
applications, except the extreme situation in which $M$ is close to $N$. 

\subsubsection{Early stopping}
We review the speed up provided by the early stopping Scheme
\ref{schemePartialSumEarlyStopOrderNeigh} to the partial sum DSOM. The
performances are reported as the ratio between the running time of the partial
sum algorithm and the running time of the early stopping algorithm.

Table \ref{tablePartialSumEarlyStoppingOrdered} summarizes the results
obtained by Algorithm \ref{algoDSOM} used with Scheme
\ref{schemePartialSumEarlyStopOrderNeigh}. As expected, the improvements
appear with high values of $M$. Moreover, the early stopping has only an
effect on the representation phase whose cost is $O(NM^2)$. If this term is
dominated by the pre-calculation phase ($O(N^2)$) the improvement will remain
unnoticed. This is why in Table \ref{tablePartialSumEarlyStoppingOrdered} the
speed up is roughly increasing with $M$ and decreasing with $N$. While in some
extreme cases, that is when $M$ is low compared to $N$ (e.g., $N=2\,000$ and
$M=49$) the ordering might be less efficient than the simple early stopping,
high values of $M$ show very good behavior. It should be noted that while this
is also observed for the real world data, it might happen in practice for the
overhead of the early stopping to be higher than reported in those
experiments.

\begin{table}[htbp]
  \centering
  \begin{tabular}{|p{10em}|c|c|c|c|c|}\hline
$N$ (data size)\newline 
$M$ (number of models) & 500 & $1\,000$ & $1\,500$ & $2\,000$ & $3\,000$ \\\hline
$49=7\times 7$         & 1.14 & 1.05 & 1    & 0.97 & 0.98\\\hline 
$100=10\times 10$      & 1.41 & 1.33 & 1.23 & 1.15 & 1.08\\\hline 
$225=15\times 15$      &      & 2.27 & 2.13 & 2    & 1.78\\\hline 
$400=20\times 20$      &      &      & 2.74 & 2.75 & 2.48\\\hline 
  \end{tabular}
  \caption{Improvement induced by early stopping with ordering}
  \label{tablePartialSumEarlyStoppingOrdered}
\end{table}

\subsubsection{Reusing earlier values}
The early stopping approach studied in the previous section reduces
the effective cost of the representation phase (whose maximal cost is
$O(NM^2)$). On the contrary, the memorization reduces the cost of the
pre-calculation phase (maximum cost of $O(N^2)$). Results presented in the
present section show that it is possible to combine both cost
reductions. 

As explained in section \ref{sectionMemory}, the behavior of Algorithm
\ref{algoDSOMSmartMemory} depends on the threshold that dictates when to use
the block update or the fine grain update. We have tested different values of
the $ratio$ parameter used in Algorithm \ref{algoDSOMSmartMemory}, from $2$
(which gives the theoretical threshold of $\frac{N}{2}$) to $9$ (threshold of
$\frac{N}{9}$). The running time depends on the chosen value, but also on $N$
and $M$. It is therefore difficult to choose an optimal universal value, but
the variability of the running times is quite small for fixed values of $N$
and $M$, especially for high values of $N$ and $M$. For $M=3\,000$ and
$M=400$, for instance, the running time of the DSOM with memory varies between
$246.3$ and $258.5$ seconds when $ratio$ varies in $\{2,3,\ldots,9\}$. Our
tests lead us to choose a ratio of $7$ for all the experiments but this
provides only a rough guideline.

Table \ref{tablePartialSumMemory} summarizes improvement factors obtained by
the DSOM with memory (Algorithm \ref{algoDSOMSmartMemory} with affectation
Scheme \ref{schemePartialSum} that does not use early stopping). As expected,
the improvement increases with $N$ as the memorization algorithm 
reduces the actual cost of the $O(N^2)$ phase. The efficiency of the algorithm
decreases with $M$ for two reasons. First point, the representation phase is
not improved by the memorization algorithm and becomes more and more important
in the global cost.  Second point, $M$ corresponds to the number of available
clusters: a big number of clusters allows more cluster modifications during the
algorithm and therefore reduces memorization opportunities. 

\begin{table}[htbp]
  \centering
  \begin{tabular}{|p{10em}|c|c|c|c|c|}\hline
$N$ (data size)\newline 
$M$ (number of models) & 500 & $1\,000$ & $1\,500$ & $2\,000$ & $3\,000$ \\\hline
$49=7\times 7$         & 1.6  & 1.92 & 2.09 & 2.02 & 2.37\\\hline 
$100=10\times 10$      & 1.2  & 1.19 & 1.26 & 1.31 & 1.53\\\hline 
$225=15\times 15$      &      & 1.18 & 1.19 & 1.21 & 1.26 \\\hline 
$400=20\times 20$      &      &      & 1.08 & 1.06 & 1.06\\\hline 
  \end{tabular}
  \caption{Improvement induced by memorization}
  \label{tablePartialSumMemory}
\end{table}

Table \ref{tablePartialSumMemoryESO} summarizes improvement factors obtained
by using the Fast DSOM which consists in Algorithm \ref{algoDSOMSmartMemory}
(Memory DSOM) with the ordered early stopping Scheme
\ref{schemePartialSumEarlyStopOrderNeigh}. Again, results reflect the
theoretical expectation. Indeed, improvements are in general much better than
those reported in Table \ref{tablePartialSumMemory}, especially for large
values of $M$. They are also better than results reported in Table
\ref{tablePartialSumEarlyStoppingOrdered}. This means that the Fast DSOM is
able to combine improvements from both memorization and early stopping. 

\begin{table}[htbp]
  \centering
  \begin{tabular}{|p{10em}|c|c|c|c|c|}\hline
$N$ (data size)\newline 
$M$ (number of models) & 500 & $1\,000$ & $1\,500$ & $2\,000$ & $3\,000$ \\\hline
$49=7\times 7$         & 1.6  & 1.92 & 2    & 2.02 & 2.39 \\\hline 
$100=10\times 10$      & 1.85 & 1.81 & 1.66 & 1.65 & 1.83\\\hline 
$225=15\times 15$      &      & 2.87 & 2.67 & 2.57 & 2.43\\\hline 
$400=20\times 20$      &      &      & 3.04 & 3.2  & 2.89\\\hline 
  \end{tabular}
  \caption{Improvement induced by memorization and ordered early stopping}
  \label{tablePartialSumMemoryESO}
\end{table}

The running times of the Fast DSOM algorithm are in fact compatible with a
real world usage for moderate data size. Running the Fast DSOM on $3\,000$
observations with a $20\times 20$ hexagonal grid takes less than 92 seconds on
the chosen hardware. The brute force DSOM algorithm needs more than $7\,150$
seconds (almost two hours) to obtain exactly the same result. The partial sum
DSOM needs approximately 264 seconds on the same data.

\subsection{Real world data}
To evaluate the proposed algorithm on real world data, we have chosen a simple
benchmark: clustering of a small English word list. We used the SCOWL
word lists \citep{SCOWL}. The smallest list in this collection corresponds to
$4\, 946$ very common English words. After removing plural forms and possessive
forms, the word list reduces to $3\, 200$ words. This is already a high value
for the DSOM algorithm, at least for its basic implementation. A stemming
procedure can be applied to the word list to reduce it even more. We have
used the Porter stemming algorithm\footnote{We have used the Java
  implementation provided by Dr. Martin Porter at \url{http://www.tartarus.org/~martin/PorterStemmer/}.} \citep{Porter80Stemming} and obtained this
way $2\, 277$ stemmed words.  

Words are compared with a normalized version of the Levenshtein distance
\citep{Levenhstein}, also called the string edit distance. The distance
between two strings $a$ and $b$ is obtained as the length of the minimum
series of elementary transformation operations that transform $a$
into $b$.  Allowed operations are replacements (replace one character by
another), insertion and suppression (in our experiments, the three operations
have the same cost).  A drawback of this distance is that it is not very
adapted to collection of words that are not uniform in term of length. Indeed
the distance between ``a'' and ``b'' is the same than the one between ``love''
and ``lover''. We have therefore used a normalized version in which the
standard string edit distance between two strings is divided by the length of
the longest string.

We used the DSOM algorithm with four different hexagonal grids, with sizes
$M=49=7\times 7$, $M=100=10\times 10$, $M=169=13\times 13$ and $M=225=15\times
15$ (bigger grids lead to a lot of empty clusters and to bad quantization). We
used $\itermax=100$ epochs and a Gaussian kernel for the neighborhood
function.  Tables \ref{tableStemWordList} and \ref{tableWordList} report the
running time in second for the brute force DSOM algorithm, for the partial sum
DSOM and for the Fast DSOM.

\begin{table}[htbp]
  \centering
  \begin{tabular}{|l|c|c|c|c|}\hline
Algorithm        & $M=49$ & $M=100$ & $M=169$ & $M=225$ \\\hline 
Brute force DSOM & 363.1  & 821     & 1456.6  & 1875.6 \\\hline 
Partial sum DSOM &  10.6  &  18.4   &   45.6  &   66.9 \\\hline 
Fast DSOM        &   5.6  &  13.9   &   29.2  &   44.1 \\\hline 
  \end{tabular}
  \caption{Running time for $2277$ stemmed English word list}
  \label{tableStemWordList}
\end{table}

\begin{table}[htbp]
  \centering
  \begin{tabular}{|l|c|c|c|c|}\hline
Algorithm        & $M=49$ & $M=100$ & $M=169$ & $M=225$ \\\hline 
Brute force DSOM & 981.8  & 2114.3  & 3739.2  & 4737.2 \\\hline 
Partial sum DSOM &  26.1  &   38.8  &   74.3  &  103.7 \\\hline 
Fast DSOM        &  14.2  &   29.2  &   49.7  &   69.5\\\hline 
  \end{tabular}
  \caption{Running time for $3\,200$ English word list}
  \label{tableWordList}
\end{table}

The obtained timings are both consistent with the theoretical model and with
results obtained in the previous section with artificial data. The exponential
model $N^\alpha M^\beta$ given in section \ref{subsubReference} gives
acceptable prediction for the running time of the brute force DSOM (NMSE is
smaller than $0.014$). The model $\delta N^2 +\tau NM^2$ proposed in the same
section gives also acceptable prediction for the partial sum DSOM running
times (NMSE is $0.010$). 

However, the improvements of the Fast DSOM over the Partial sum DSOM are not
as important as with the artificial data (the improvement factor is between
$1.3$ and $1.9$), mostly because the effect of the early stopping are reduced:
despite the ordering, early stopping does not happen as frequently as for the
artificial data set. Nevertheless, it still appears clearly that the Fast DSOM
algorithm should always be used in practice, especially because the results
are strictly identical to those obtained with the brute force DSOM.

\section{Conclusion}
We have proposed in this paper a new implementation method for the DSOM, an
adaptation of Kohonen's Self Organizing Map to dissimilarity data. The cost of
an epoch of the standard DSOM algorithm is proportional to $N^2M+NM^2$,
where $N$ is the number of observations and $M$ the number of models. For our
algorithm, the cost of an epoch is proportional to $N^2+NM^2$. As $M$ is
in general much smaller than $N$, this induces a strong reduction in the
running time of the algorithm. Moreover, we have introduced additional
optimizations that reduce the actual cost of the algorithm both for
the $N^2$ part (a memorization method) and for the $NM^2$ part (an early
stopping strategy associated to a specific ordering of the calculation).

We have validated the proposed implementation on both artificial and real
world data. Experiments allowed us to verify the adequacy of the theoretical
model for describing the behavior of the algorithm. They also showed that the
additional optimizations introduce no overhead and divide the actual
running time by up to 3, under favorable conditions. 

The reduction in running time induced by all the proposed modifications are so
important that they permits to use the DSOM algorithm with a large number of
observations on current personal computers. For a data set with $3\, 000$
observations, the algorithm can converge in less than two minutes,
whereas the basic implementation of the DSOM would run for almost two
hours. Moreover, the proposed optimizations don't modify at all the results
produced by the algorithm which are strictly identical to the ones that would
be obtained with the basic DSOM implementation. 

\section*{Acknowledgements}
The authors thank the anonymous referees for their valuable suggestions that
helped to improve this paper. 

\bibliographystyle{elsart-harv}
\bibliography{biblio}

\end{document}